\documentclass[journal,hideappendix]{vgtc}        

\onlineid{1360}

\vgtccategory{Research}

\vgtcpapertype{algorithm/technique}

\title{Sports Analysis and VR Viewing System Based on Player Tracking and Pose Estimation with Multimodal and Multiview Sensors}

\author{%
  \authororcid{Wenxuan Guo$^*$}{0009-0007-4823-1587},
  \authororcid{Zhiyu Pan$^*$}{0009-0000-6721-4482}, \authororcid{Ziheng Xi}{0009-0008-7007-8803}, \authororcid{Alapati Tuerxun}{0009-0004-5513-4783}, \authororcid{Jianjiang Feng$^\dag$}{0000-0003-4971-6707}  and 
  \authororcid{Jie Zhou}{0000-0001-7701-234X}
}

\authorfooter{
  \item
  	Wenxuan Guo, Zhiyu Pan, Ziheng Xi, Alapati Tuerxun, Jianjiang Feng and Jie Zhou are with Department of Automation, Tsinghua University.
   
  	E-mail: \{gwx22, pzy20, xizh21, alpttex19\}@mails.tsinghua.edu.cn, \{jfeng, jzhou\}@tsinghua.edu.cn
   \item $^*$: Equal contribution.
   \item $^\dag$: Corresponding author.
}

\abstract{%
Sports analysis and viewing play a pivotal role in the current sports domain, offering significant value not only to coaches and athletes but also to fans and the media. In recent years, the rapid development of virtual reality (VR) and augmented reality (AR) technologies have introduced a new platform for watching games. Visualization of sports competitions in VR/AR represents a revolutionary technology, providing audiences with a novel immersive viewing experience. However, there is still a lack of related research in this area. In this work, we present for the first time a comprehensive system for sports competition analysis and real-time visualization on VR/AR platforms. First, we utilize multiview LiDARs and cameras to collect multimodal game data. Subsequently, we propose a framework for multi-player tracking and pose estimation based on a limited amount of supervised data, which extracts precise player positions and movements from point clouds and images. Moreover, we perform avatar modeling of players to obtain their 3D models. Ultimately, using these 3D player data, we conduct competition analysis and real-time visualization on VR/AR. Extensive quantitative experiments demonstrate the accuracy and robustness of our multi-player tracking and pose estimation framework. The visualization results showcase the immense potential of our sports visualization system on the domain of watching games on VR/AR devices. The multimodal competition dataset we collected and all related code will be released soon. 
}


\teaser{
  \centering
  \includegraphics[width=\linewidth]{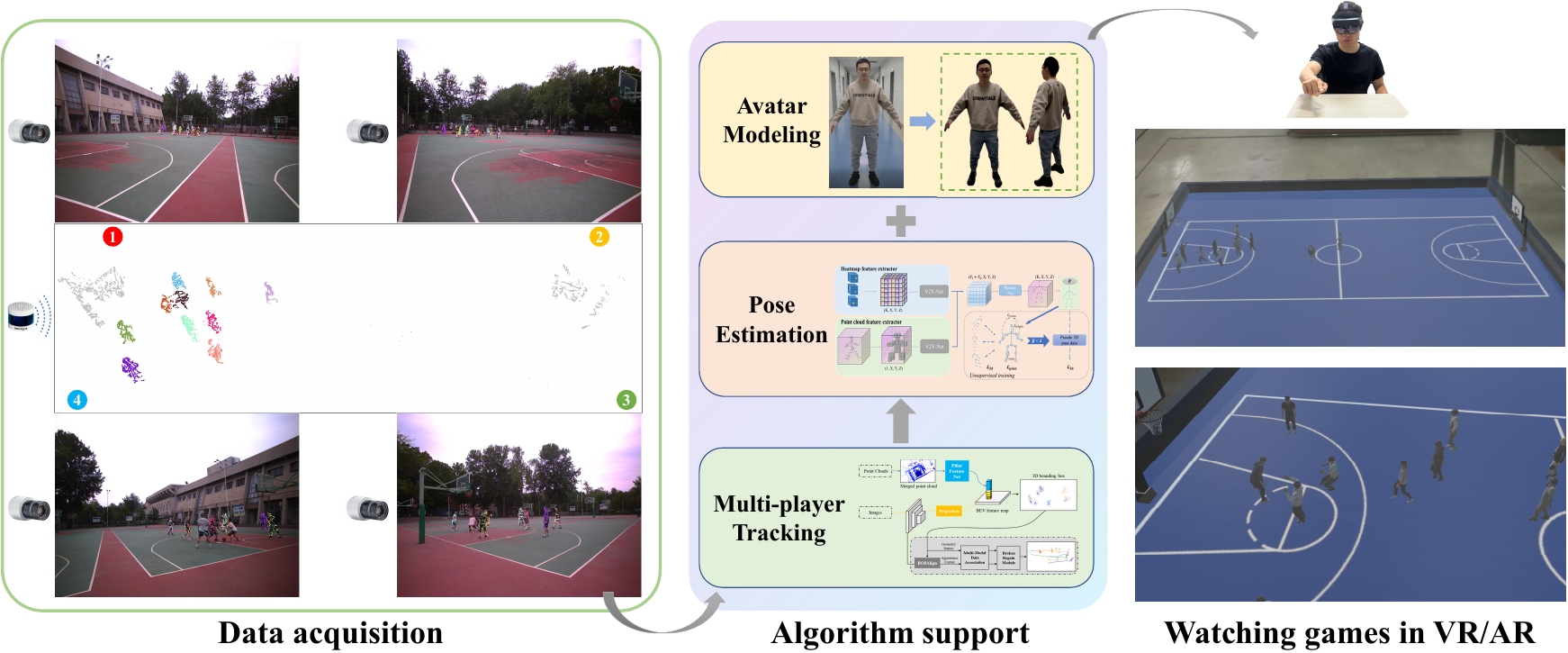}
  \caption{The overview of our proposed sports analysis and VR viewing system. Firstly, game data is collected by multimodal and multiview sensors, including LiDAR and cameras. Subsequently, multi-object tracking and pose estimation algorithms are employed to perceive the positions and movements of players, while avatar modeling algorithm are used for creating 3D drivable models of the them. Finally, we drive these avatar models based on the position and pose results, visualizing them in virtual space. In this way, audiences can watch the game on VR/AR devices.
  }
  \label{fig:intro}
}

\graphicspath{{figs/}{figures/}{pictures/}{images/}{./}} 

\usepackage{tabu}                      
\usepackage{booktabs}                  
\usepackage{lipsum}                    
\usepackage{mwe}                       

\usepackage{amsmath, bbm, amssymb, multirow}
\usepackage{tabularx}
\usepackage{float}
\begin{document}


\firstsection{Introduction}

\maketitle

With the rapid development of technology, the sports industry is undergoing an unprecedented transformation. The close integration of technology and sports has opened up new areas of research and practice, demonstrating immense potential and value. This fusion leverages the latest technological advancements with the aim of comprehensively enhancing the performance within the sports sector. This includes, but is not limited to, improving training methodologies, increasing the spectacle of competitions, and enhancing athlete safety. In this process, sports analysis and game visualization technologies play a crucial role.

Sports competition analysis, by collecting and analyzing vast amounts of game data, assists coaching teams and audiences in gaining an in-depth understanding of key moments in matches and athletes' performances. This enables the formulation of more scientific and precise training plans and competition strategies. Additionally, through biomechanical analysis and motion capture technology, coaches and scientists can meticulously analyze athletes' movement patterns. This helps identify inefficient or potentially injurious actions, thereby optimizing athletes' training regimes and techniques.

The integration of sports competitions with Virtual Reality (VR) or Augmented Reality (AR) technologies opens a new chapter in the spectator experience. VR/AR technologies enable spectators to feel as if they are on-site, watching the game from various angles and distances, and choosing to follow the perspective of a specific athlete, which significantly enhances the immersion of watching a game. For fans unable to attend the games, VR/AR viewing offers an excellent solution. Additionally, viewers can interact with other spectators or elements in the virtual environment, providing a more personalized and interactive viewing experience. With VR/AR products~\cite{AppleVisionPro2024,OculusQuest2024,HoloLens2024} becoming increasingly common and their performance improving, the foundation for VR/AR viewing has been established. However, current VR/AR viewing methods~\cite{OculusQuest2024,AppleVisionPro2024} generally require a large array of cameras to collect game data, leading to high costs and making it difficult to apply to regular competitions. Furthermore, they require high computational costs for rendering, making it difficult to broadcast events in real-time, limiting their use to game replays. Therefore, there is a pressing need to explore new low-cost and real-time VR/AR viewing frameworks.

In this paper, we approach the challenge of watching games on VR/AR devices with a novel perspective, achieving analysis of game data meanwhile. 
Firstly, game data is collected by multimodal and multiview sensors, including LiDAR and cameras. Subsequently, multi-object tracking and pose estimation algorithms are employed to perceive the positions and movements of players, while avatar modeling algorithm are used for creating 3D drivable models of the them. Finally, we drive these avatar models based on the position and pose results, visualizing them in virtual space. Furthermore, we construct a new multimodal dataset for evaluating our method. The dataset comprises nearly 11,000 frames captured from two basketball games in various real-world scenarios. 

3D multi-object detection and tracking is essential for visual surveillance and player tracking~\cite{soccer_survey,epfl}. Most of recent multi-object tracking methods~\cite{center_track,Chiu,GNN3DMOT} follow the tracking-by-detection paradigm, in which objects are first obtained by detection and then linked to previous trajectories by data association. The resulting trajectories may then be utilized for sports analysis like running speed and total distance covered~\cite{linke2020football}. Traditional player tracking systems typically employ multiview RGB cameras for observing the scene~\cite{epfl,ksp,graph_mcmt,VoxelTrack}. However, poor lighting, limited ability to identify foreground, similar appearance, and inaccurate depth estimations have an impact on how well these methods perform. In recent yeas, LiDAR-based systems~\cite{FaF,DBLP:conf/iros/ab3Dmot,Chiu,PointTrackNet} have developed rapidly and become increasingly cheaper, enabling precise 3D object detection and tracking in autonomous driving scenes. Since point clouds provide precise 3D information, robust tracking trajectories may be produced based on 3D detection results. However, due to the lack of appearance information, LiDAR-based systems are prone to tracking errors due to poor detection performance for extremely close objects, which is very common in sport games.

In this work, we propose an online tracking method that fuses multiview LiDARs and cameras. In order to improve the reliability and accuracy of dynamic object perception, multimodal sensors can be employed to provide richer clues and hence reduce detection and tracking errors. To our best knowledge, this is the first multi-player tracking research taking multiview RGB and point cloud as input.

Human pose estimation is a fundamental task in computer vision and has been widely applied in various fields, such as human-computer interaction, human-robot interaction, human activity recognition, etc. Specifically, multiview image datasets \cite{human36m,panoptic,3dps,3dpw} allow more precise 3D human pose estimation compared to single-view ones \cite{mpii,pose2seg,COCO,lin2014microsoft}, due to the ability of multiple views to capture 3D information from epipolar geometry. As technology advances, researchers \cite{2019learnable,2020voxelpose,2021multi,2021direct,2021tessetrack} have achieved promising results on these multiview images datasets. However, practical scenarios are more challenging than existing public datasets, with diverse human motions, severe occlusions, and large scene. Combining different sensors such as LiDAR and RGB camera has the potential of boosting the accuracy of 3D human pose estimation performance in multiple situations. It is essential to consider integrating the information from sensors of various types and views, and fusing them into a cohesive representation which can significantly facilitate the accuracy of 3D human pose estimation.

In this paper, we propose PointVoxel, a 3D human pose estimation pipeline. Volumetric representation is an effective and reasonable architecture to fuse RGB and point cloud. Researchers \cite{2019learnable,2020voxelpose,2021tessetrack,2022faster} have shown the effectiveness of voxel-based method. Voxel-based architecture can naturally model a space's geometric characteristics. Besides, it is straightforward to either map the point cloud information or back-project the 2D information into the 3D volumetric representation. As to multimodal features fusion \cite{hansen2019fusing,gerats20223d}, we refer to the work BEVFusion \cite{BEVFusion} and unify different modalities into a volumetric space that reserve each modality's geometric characteristics. 
Manually annotating or capturing 3D poses for multiple individuals in large scenes is challenging, expensive, and hard to ensure generalizability to new scenarios. To achieve better results on challenging scenarios without pose labels, we adopt the approach of pretraining on a synthetic dataset first and then unsupervised domain adaption training on the target dataset.

In the film and gaming industries, the creation and animation of virtual characters is a complex and detailed process, with avatar modeling being one of the key components. Avatar modeling requires the models to be visually appealing and capable of simulating complex actions. This process involves several steps, including concept design, 3D modeling, texture mapping, rigging, and body animation. In recent years, to perform 3D modeling of real human, some works focus on using multiview photos or videos for real human avatar modeling. Our VR/AR viewing framework follows the modeling and driving processes in the animation production field. It involves pre-modeling the athletes in the competition using real human avatar modeling techniques, and driving the virtual athletes based on their positions and poses obtained through multi-player tracking and pose estimation. We utilize the software Unity to achieve 3D reconstruction and rendering of the competition, deploying it on VR/AR terminals to provide audiences with an immersive viewing experience.

To summarize, our contributions are as follows:
\begin{enumerate}
    \item We propose a novel sports competition analysis and VR viewing system that boasts low cost and real-time performance. It utilizes multimodal and multiview data from LiDARs and cameras, and follows a process including data collection, multi-player tracking, pose estimation, player avator modeling, and actuation.
    \item We propose a method for multi-player tracking and pose estimation that integrates multimodal and multiview data, aiming to achieve precise 3D results. For pose estimation, we present a new training strategy without using high-cost manual annotations.
    \item We construct a novel multimodal and multiview dataset of sport competitions to facilitate development in this area. Extensive experiments demonstrate the accuracy of our multi-player tracking and pose estimation method, as well as the significant potential of our VR/AR viewing system. 
\end{enumerate}

\section{Related Work}
\subsection{Sports Visualization in VR/AR}
Virtual Reality (VR) and Augmented Reality (AR) technologies offer a new platform for viewing sports competitions, providing the global audiences with immersive and interactive participation experiences. 

In the terms of VR, devices like Oculus Quest~\cite{OculusQuest2024} and Pico Neo~\cite{PicoNeo32024} allow users to enjoy a 360-degree view of the action, making them feel as if they are sitting in the stadium. AR devices~\cite{HoloLens2024,AppleVisionPro2024}, on the other hand, enriches the live viewing experience by overlaying digital information onto the real-world environment. Applications on AR allow users to see player stats, game dynamics, and other relevant information in real-time while watching the game, either on-site or from a remote location. This not only makes the experience more engaging but also helps fans understand the game better. Moreover, the social aspect of watching sports could be enhanced through VR/AR technologies. Platforms such as Oculus Quest~\cite{OculusQuest2024} and social features in apps like LiveLike~\cite{LiveLike2024} enable fans to watch games together virtually, regardless of their physical location. This can create a sense of community and shared experience that was previously only possible by attending the games in person.

Currently, the NBA and FIFA, as two of the world's leading sports event organizations, have adopted a series of innovative measures in VR broadcasting and viewing to enhance the spectator experience and interactivity. For certain games, they offer VR viewing applications, allowing users to watch live broadcasts and replays of highlights through head-mounted devices. This viewing method offers the possibility to watch the game from different angles, including courtside and aerial views, bringing an unprecedented level of immersion and a sense of being at the venue to fans. However, existing solutions generally require a large array of cameras to collect game data, leading to high costs that make it difficult to apply to regular competitions. Additionally, they require significant computational resources for rendering, making it challenging to broadcast events in real time. Currently, they only support viewing from specific locations and within a limited range of angles, and do not allow for more complex viewing modes, such as choosing to follow the perspective of a specific athlete. 
Therefore, our work explores a low-cost, highly operable, and real-time framework for viewing sports competitions through VR/AR.

\subsection{Multi-object Tracking}
Recent Multi-object tracking (MOT) systems mostly follow the tracking-by-detection paradigm, in which objects are first obtained by a detector and then linked to previous trajectories by data association.
DeepSort~\cite{Deepsort} uses Kalman filter to update the trajectories and solves data association as the bipartite matching problem using the Hungarian algorithm. CenterTrack~\cite{CenterTrack}, on the other hand, directly regresses the offset to obtain the data association across frames. These methods typically use deep features of images for cross-frame association, lacking accurate 3D geometry and location information.

For 3D MOT in the area of visual surveillance, multi-cameras~\cite{epfl,ksp,graph_mcmt,VoxelTrack} are often used to calculate the 3D location of the objects to obtain 3D tracking results. KSP~\cite{ksp} first obtains the detection by background subtraction, projects it onto the ground to create a probability occupancy map, and finally uses the K-shortest paths optimization to obtain offline tracking results. Some methods\cite{MMSportsVideo,LMGP} followed it by using a 2D detector to improve the detection results. In the field of sports game analysis, Zhang \textit{et al.}~\cite{MMSportsVideo} propose a robust multi-camera player tracking framework by improving the KSP algorithm using player identification. In addition, the majority of player tracking datasets are captured by multiview cameras, such as APIDIS dataset\cite{apidis} and STU dataset\cite{MMSportsVideo}. However, the 3D depth obtained by epoipolar geometry is not accurate enough for robust association and tracking.

As for autonomous driving scenes, LiDAR-based approaches~\cite{FaF,DBLP:conf/iros/ab3Dmot,Chiu,center_track,Complexer} are dominant. Weng \textit{et al.}~\cite{DBLP:conf/iros/ab3Dmot} propose a simple and robust baseline based on point clouds, using 3D Intersection over Union (IoU) as the data association. Chiu \textit{et al.}~\cite{Chiu} use the Mahalanobis distance instead of IoU to improve the association accuracy. CenterPoint~\cite{center_track} is an extension of CenterTrack~\cite{CenterTrack}, which solves the data association by regressing the offset of the object. Most of these methods rely on 3D point clouds and lack sufficient appearance information, which may lead to tracking errors for very close objects.

\subsection{3D Human Pose Estimation}
    \noindent
    \textbf{Image-based.}
    Basic 3D pose estimation method is in two stage that estimates the 2D pose first and then lifts it into 3D space  
    \cite{2019fast,2019learnable,2020voxelpose,2021multi,2021graph,2021tessetrack}. As to multi-person setting, some methods \cite{2019fast,2021multi} match pedestrians from 
    different views and then locate them through 2D pose similarity. But they are not robust to inaccurate 2D pose results. Zhang et al. \cite{2021direct} 
    directly utilizes the 2D images as inputs and regresses the 3D pose by transformer architecture. However, their training process is time-consuming. Some voxel-based methods \cite{2020voxelpose, 2021tessetrack} locate each person in a 3D volumetric space and estimate 3D poses. 
    Such voxel-based methods greatly improve the precision of 3D human pose estimation. However, they are not suitable for the large scene because of 
    large computational cost when detecting people.
 
    \noindent
    \textbf{Point cloud-based.}
    3D pose estimation methods focus on single-view depth map \cite{garau2021deca,haque2016towards,guo2017towards,martinez2020residual} at the beginning. They treat the depth map as 2D 
    information with depth value. On the other hand, a few of methods back-project depth map into 3D space as dense 3D point cloud and use
    Pointnet network to deal with \cite{zhang2021sequential,bekhtaoui2020view}. Moon et al. \cite{moon2018v2v} propose a single-person 
    pose estimation approach which treats depth map as point cloud and fill it in volumetric space.  
    Bekhtaoui et al. \cite{bekhtaoui2020view} use Pointnet related approach to detect and estimate 3D human pose. 
    Recently, Li et al. \cite{lidarcap} use sparse point cloud scanned from LiDAR to estimate single-person 3D human pose. However, sparse 
    point cloud cannot provide enough information for accurate 3D pose estimation. Multimodal fusion is helpful for concise 3D human poses' perceiving.
 
    \noindent
    \textbf{Multimodal based.}
    Many works have been introduced for RGBD human pose estimation \cite{bashirov2021real,zheng2022multi,hansen2019fusing,gerats20223d}. 
    Multimodal information can not only help detect the person 
    but also guarantee the accuracy of 3D human pose. Lately, some researchers \cite{cong2022weakly,lidarcap} introduce camera-LiDAR setting datasets 
    for outdoors and large scene. As to the multimodal fusion methods, some approaches \cite{zheng2022multi,bekhtaoui2020view, hansen2019fusing} 
    fuse modalities in point-level by attaching extracted 2D features to every 3D point. And a few methods \cite{piergiovanni20214d,liang2018deep} fuse modalities in 
    feature-level strategy. They do not unify the 2D and 3D information in a unified interpretable space, and fuse them in an 
    uninterpretable feature space instead.
 
    Our approach utilizes multimodal information for human detection and 3D human pose estimation. 
    Besides, we use volumetric representation representing real-word space to unify the 2D and 3D information.

\begin{figure*}
    \centering
    \includegraphics[width=0.8\linewidth]{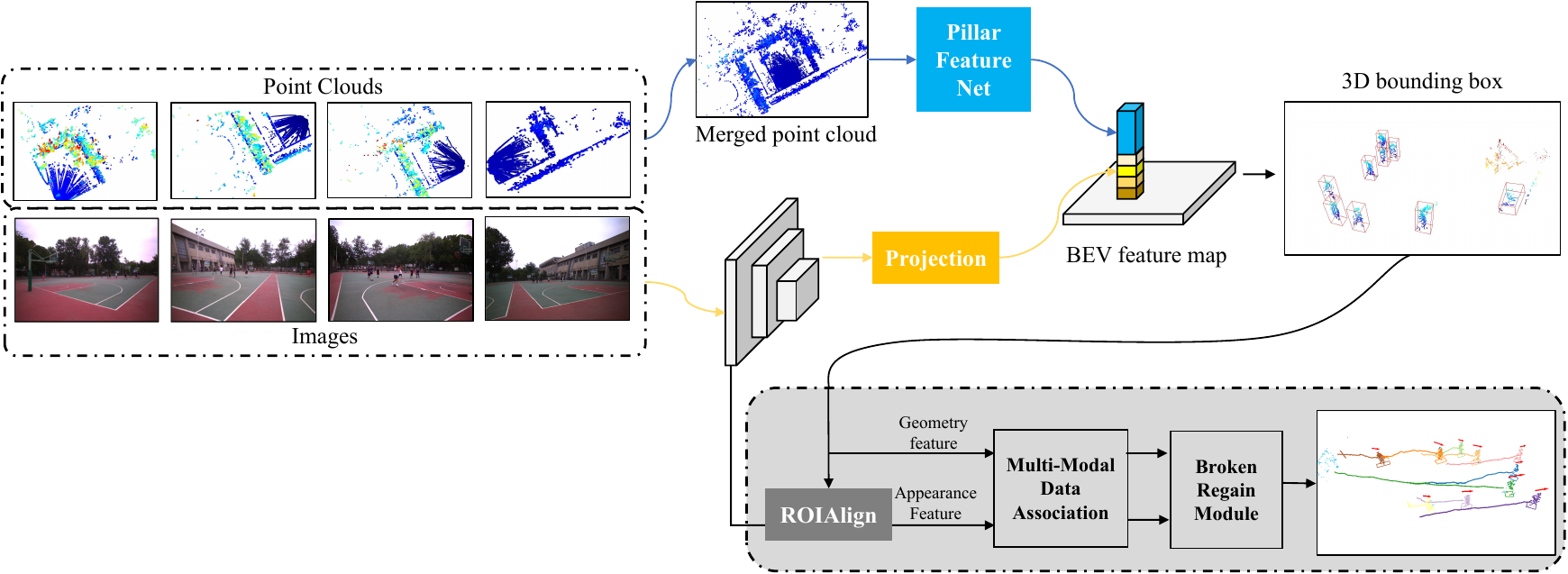}
    \caption{Overview of our multi-player tracking method. We extract features in images and point clouds and fuse them in BEV for multimodal detection results. Both geometry information and appearance information are leveraged for long-term object tracking.}
    \label{fig:method}
\end{figure*}
\section{Method}

Our proposed sports analysis and VR viewing system utilizes multimodal and multiview point clouds and images as input, ultimately enabling watching games on VR/AR devices while providing accurate player statistics. In the following, we introduce the key technologies within the system. In Sec.~\ref{sec:3dmot}, we present our proposed method for multi-player tracking. In Sec.~\ref{pose_esti}, we introduce our pose estimation method along with an unsupervised training scheme. In Sec.~\ref{avatar}, we describe avatar modeling and the VR/AR visualization methods. 

\subsection{Multiview Multimodal 3D Tracking}\label{sec:3dmot}
We propose a 3D online multi-object tracking approach, which improves
reliability and accuracy via modality fusion. As shown in Fig.~\ref{fig:method}, our framework follows the widely employed tracking-by-detection paradigm. In the following, we first describe the multimodal detection stage in Sec.~\ref{sec:detect}, which takes as input multiview images and merged multiview point clouds
and outputs the 3D bounding boxes. The data association stage is presented in
Sec.~\ref{sec:track} which links the detection results to the previous trajectories. 

\subsubsection{\textbf{Multimodal Detection}}\label{sec:detect}

Our multimodal detection method consists of two streams: an image processing network and a point cloud processing network, designed to extract modal-specific feature maps. The multimodal feature map is obtained by combining the two modal-specific feature maps based on spatial position in bird's eye view (BEV). Finally, a detection head is employed to obtain accurate 3D object detection results.

\noindent\textbf{Point Cloud Processing Network.}
The point clouds from all single views are spatio-temporally aligned and fused to obtain a multiview dense point cloud. The feature encoder and the backbone make up the two components of the point cloud processing network. Pillar Feature Net~\cite{pointpillars} is adopted by the feature encoder to transform point clouds into pseudo images, while backbone is used to extract high-level features of the pseudo images.

Firstly, the point cloud is assigned to multiple grids in the X-Y plane, so that a frame of point cloud data can be encoded into a dense tensor of dimension $(D,P,N)$, where $D$ stands for the feature dimension, $P$ for the number of grids, and $N$ for the number of point clouds contained within each grid. We then utilized a simple PointNet~\cite{pointnet} to create a tensor of dimension $(C_p,P)$. Finally, the pseudo image of $(C,H_p,W_p)$ is generated by the scatter operator.

The backbone is made up of two sub-networks. A top-down sub-network is used to extract features on increasingly small spatial resolutions, and the other sub-network is responsible for up-sampling the features to the same dimensional size by a deconvolution operation. The final output is a combination of all features from different resolutions.

\noindent\textbf{Image Processing Network.}
The input to the image processing network is images captured by cameras with different perspectives. The input images are processed simultaneously by a convolutional network to extract features, which are then projected to BEV space for feature aggregation. We choose the lightweight ResNet18~\cite{resnet} as the backbone because of its efficiency and capability of feature extraction. The convolutional network computes the $C_i$-channel feature maps of each of the $M$ input images with shared weights. Before projection, the $M$ feature maps are resized to a fixed size $(H_i, W_i)$, where $H_i$ and $W_i$ denote the height and width of the feature maps.

Multiview information aggregation is performed by the projection transformation. Assuming that the ground is horizontal, we can get the correspondence between the pixels of multiview images and the coordinates of ground point with the extrinsic and intrinsic parameters. To enhance the perception of spatial location, multiple projected feature maps and X-Y coordinate maps are concatenated to obtain the multiview image feature map of $M\times C_i+2$ channels.

\noindent\textbf{Detection Head and Loss.}
The image and point cloud feature maps are concatenated according to spatial locations in BEV to obtain multimodal feature maps, and then a detection head is utilized to obtain 3D object detection results.

Considering the simplicity and computational complexity, we use Single Shot Detector (SSD)~\cite{SSD} as the detection head, which using 2D IoU to match the detection results with the ground truth during training. We apply the same loss functions introduced in SECOND~\cite{second}.

\subsubsection{\textbf{Multimodal MOT}}\label{sec:track}
Given the 3D object detection results of the current frame, previous trajectories are associated by fusing geometry information and appearance information. As for the geometry distance metric, we employ the 3D distance intersection-over-union (3D DIoU) between the detection bounding boxes and the tracking bounding boxes. The 3D object detection boxes are projected onto the input multiview 2D camera planes to obtain the image patches, which are used to extract the appearance features of objects. And then we take into account the cosine distance of appearance features between the detections and the tracks as the appearance distance metric. We design a linear combination of the geometry distance and appearance feature distance to aggregate the two affinity matrices, which is passed to the Hungarian algorithm for data association and Kalman filter for prediction.

\noindent\textbf{Multimodal Data Association.}
To associate the detections with the tracks, we calculate the 3D DIoU~\cite{DBLP:conf/aaai/DIoU} between every detection-track pair. Assume that the detections $D^t = \{d^t_i|i=1,...,N\}$ generated by the multimodal detector, and the tracks $G^{t} = \{g^t_i|i=1,...,M\}$ predicted by the tracking module of $t-1$ frame, each $d^t_i, g^t_i$ is a 3D bounding box denoted by centers, dimensions and the rotation angle. Considering the penalty term of center distance, we employ the 3D DIoU to avoid the collapse of the normal IoU for non-overlapping cases, which is defined as follows:
\begin{equation}
    G(d^t_i, g^t_i)=0.5\times (\mathit{IoU_{3D}}-\frac{\rho^2(c_d,c_g)}{\rho^2(d,g)}+1)
\end{equation}
where $c_d,c_g$ are the center of two bounding boxes $(d^t_i, g^t_i)$ and $\rho(d,g)$ is the distance of their farthest corners. With the penalty term, two bounding boxes far apart will get a lower 3D DIoU score, which ranges from zero to one. We apply the 3D DIoU score for all detection-track pairs to obtain the geometry affinity matrix $S_{ge} \in R^{M\times N}$.

Despite obtained 3D bounding boxes have accurate 3D location information, using only geometry information is prone to errors since it's hard to distinguish the close objects from aggregated point clouds. Therefore, we also take into account the appearance information for association metric. We project the 3D detection bounding boxes onto the input multiview camera planes to obtain the image patches. It should be noted that the projection of the 3D bounding boxes onto the 2D images can cause occlusion problems. We determine the level of occlusions for each view separately, by calculating the 2D IoU score of each detection box. If the IoU of the two bounding boxes is greater than the threshold, we consider the object with larger depth to be occluded by the close object, which indicates that the 2D appearance feature of the occluded object is unreliable.
For each valid projected bounding box, we extract a $512 \times 25 \times 25$ image feature from the RoIAlign~\cite{maskrcnn} feature of the image processing stream in Sec.~\ref{sec:detect}, which is max-pooled to a $512$ dimensional tensor as the appearance feature. 

\begin{figure*}[]
\centering
\includegraphics[width=0.85\linewidth]{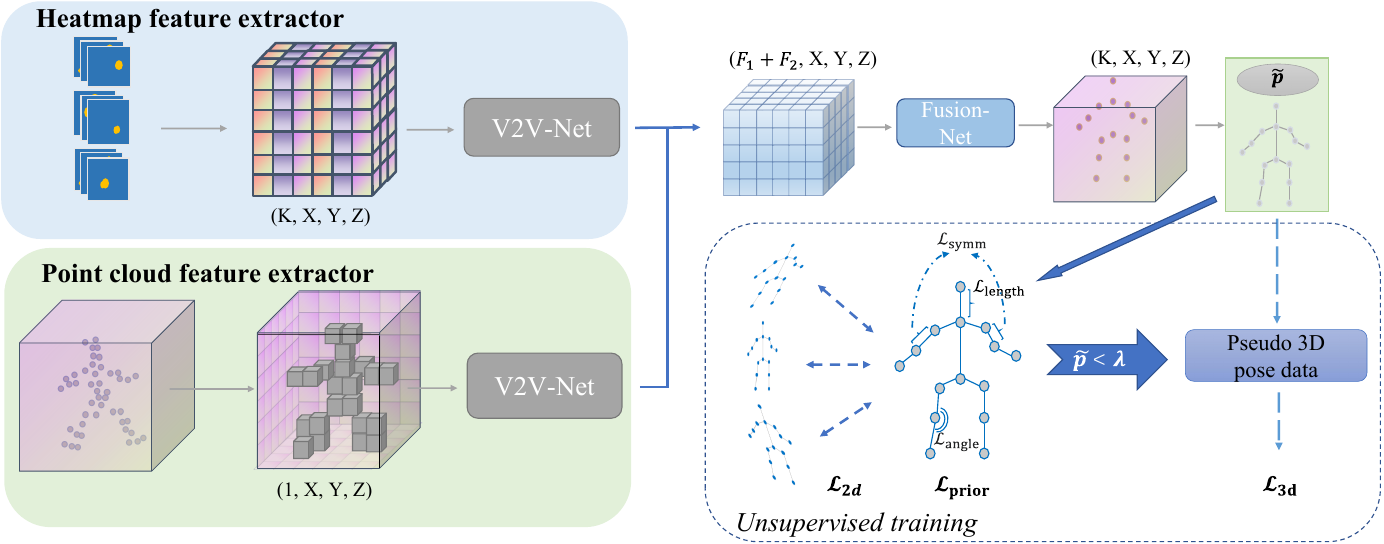}
\caption{The detailed structure of PointVoxel in 3D human pose estimation and its unsupervised training strategy. Dashed arrows indicate 
calculating corresponding loss. \includegraphics[height=0.2cm]{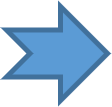} represents conditional flow which means the pose will be the 
pseudo 3D label for the next epoch if required condition is met.} 
\label{fig:pointvoxel}
\end{figure*}

For each object $d^t_i$, our method extracts its appearance features $F^t_i= \{f^t_{i,j}|j =1,...,C\}$ from $C$ valid views. Meanwhile, the appearance features of tracks $g^{t}_m$ is obtained from \textit{feature memory bank}, which consists of $K$ feature tensors of all valid image patches in the past $l$ frames. The cosine similarity of feature tensors is used as the appearance distance metric, which is defined as follows:
\begin{equation}
    {A(d^t_{i,j}, g^{t}_{m,n})}= \frac{f^t_{i,j}\cdot f^t_{m,n}}{\|f^t_{i,j}\|\|f^t_{m,n}\|}
\end{equation}
where $f^t_{i,j}$ are the appearance features of the $i$th detection in the $j$th view, and $f^t_{m,n}$ are the $n$th appearance feature in \textit{feature memory bank} of the $m$th track.  For each detection-track pair, a cosine similarity matrix of size $C\times K$ can be obtained, and then we take the top $k$ largest cosine similarity values of each row as the valid matches for each view. The valid cosine similarity scores $C\times k$ are eventually averaged as the final appearance similarity of this detection-track pair. If the valid appearance features of the track is less than $k$ or there is no valid feature for this detection, which means that the appearance information is not reliable, we use the 3D DIoU score as the default value. Finally, we obtain the appearance affinity matrix as $S_{ap} \in R^{M\times N}$.

With the geometry affinity matrix $S_{ge}$ and appearance affinity matrix $S_{ap}$, we design a linear combination of the two affinity matrices to aggregate the two affinity matrices, which is passed to the Hungarian algorithm for data association and Kalman filter for prediction. The final affinity matrix $S$ is defined as follows:
\begin{equation}
    S = \alpha S_{ge} + (1-\alpha)S_{ap}
\end{equation}
where $\alpha$ is the proportion of images and point clouds contribution to the data association. After the data association at $t$ frame, the obtained appearance features of the object $d^t_i$ are added to the \textit{feature memory bank} of $g^{t}_m$ for the next frame.

\noindent\textbf{Trajectory Regain Module.}
This module is designed to regain the broken trajectories, which are caused when objects are extremely close or out of the field. After the objects are re-detected, the broken trajectories should be regained for a long-term tracking. However, traditional tracking systems~\cite{DBLP:conf/iros/ab3Dmot,center_track} tend to treat them as new objects lack of appearance information.

To regain the broken trajectories, we manage all unmatched detections and unmatched trajectories using appearance features under geometry constraints. On the one hand, we consider all unmatched trajectories as probable existed objects which are missed due to extremely close or out of the field. Instead of deleting the potential broken trajectories, we keep tracking them and retain their appearance features in the memory bank. The location of the potential broken trajectories is predicted by Kalman filter until the objects leave the field. 

On the other hand, we consider all unmatched detections as the potential objects of the broken trajectories before. We calculate the regain score between each unmatched detection $d^t_i$ and each broken trajectory $g^t_j $ under the geometry constraints. In sports competitions, like basketball and football games, the 3D location of the player does not change abruptly, therefore players won't suddenly appear or disappear from inside the court. We construct geometry constraints by assuming the number of objects in the local area is constant, that is to say, the number of objects in the local area is the same as the number of objects in the historical frames. For the unmatched trajectory $g^t_j$, if the object is moving out of the area by the edge, we consider the unmatched detection enters from the same edge to meet the geometry constraints. If the broken trajectory is in the area, we consider a neighboring area as the local area, whose radius is improving by the time since the trajectory has been broken, and the corresponding unmatched detection should be in the local area to meet the geometry constraints.

The regain score of inner tracks is defined by exploiting both geometry and appearance information: $s = \alpha G(d, g) + (1-\alpha){A(d,g)}$. As for the objects moving out of the field by the edge, the regain score is considered by the appearance similarity $A(d,g)$.
The score is set to $-1$ if the constraints are not met. If the maximum score $S(d^t_i,g^t_j)$ is greater than the threshold, the detection and the trajectory are considered to belong to the same object, and thus the broken trajectory is maintained and interpolated to improve the continuity of the long-term object tracking.

\subsection{Multi-player Pose Estimation} \label{pose_esti}
    The proposed method is composed of two parts: 3D human pose estimation and unsupervised domain adaption training, as shown in Fig.~\ref{fig:pointvoxel}.  
    Section \ref{pointvoxel} introduces the approach of the top-down 
    manner of pose estimation. We introduce the details about voxel-based multimodal fusion. Section \ref{unsup} illustrates how we generate the synthetic dataset to 
    support the training of multimodal pose estimation model, and how to combine entropy with the human prior loss to realize domain shift. 
 \subsubsection{\textbf{PointVoxel}} \label{pointvoxel}
    Similar to \cite{2020voxelpose, 2019fast,2021tessetrack}, we adopt the top-down manner to estimate the 3D pose. Through the multi-player tracking method described in Sec.~\ref{sec:3dmot}, we obtain long-term 3D and 2D bounding boxes for each player.
    At first, we can define a volumetric space centered at the poincloud's centroid whose size is consistent 
    with detected bounding boxes', and discretize it into an $X \times Y \times Z$ resolution. So that we can fill each voxel according to each point cloud's 
    coordinate. In our case, we set value $1$ to the voxel containing poincloud. Then we can get the point cloud-related feature ($F_1 \times X \times Y \times Z$) via 
    3D convolution backbone V2V-Net \cite{moon2018v2v}.
    
    Thanks to the development of 2D pose estimation methods \cite{alphapose,vitpose}, relatively
    accurate corresponding 2D pose can be predicted without extra training. Hence, we can get each view's 2D pose heatmap via Gaussian blur. 
    We apply the same projecting way as \cite{2020voxelpose} to calculate each voxel's 
    information, and apply V2V-Net to extract the rgb-related features ($F_2 \times X \times Y \times Z$) in 3D space. Then, we concatenate the two modal features 
    in the same 3D space for modality fusion and get the final 3D human pose heatmap via a 3D convolution head Fusion-Net. In order to 
    mitigate quantization error, we adopt \textit{Soft-argmax} to calculate each joints' 3D coordinate $J^k$ through 3D heatmaps and minimize 
    the $L_1$ loss with the ground truth $J_*^k$: 
    \begin{equation}
    \begin{aligned}
       \label{poseloss}
       \mathcal{L}_{\text{pose}} = \sum_{k=1}^{K} \left \Vert J^k - J_*^k \right \Vert_1
    \end{aligned}
    \end{equation}
    where $K$ is the number of joints.
 
 \subsubsection{\textbf{Unsupervised Domain Adaptation}} \label{unsup}
    Due to the lack of annotated multiview multimodal 3D human pose dataset, we choose to generate a synthetic dataset to assist training. 
    SyncHuman we designed can create 3D dynamic scenes of multiple persons with accurate labels. Besides, we develop an efficient unsupervised domain adaption means 
    by designing a loss function to transfer the pretrained model from synthetic dataset to real-world dataset.  
    
 \noindent\textbf{SyncHuman.} \label{sync}
    We develop the synthetic system based on Unity Engine. Due to its flexibility and productivity, we can formulate a scene with a specific size and background as 
    per our specifications. Concerning sensors' aspect, we implement Unity's built-in camera to obtain RGB images, and extract the depth information from the GPU's depth buffer with custom shaders used in the rendering pipeline.  
    Furthermore, we can attain the colored point cloud by sampling the depth and RGB images. For the sake of imitating the scanning process of LiDAR, we need to sample 
    points along to a scanning function of time. Currently, we consider Livox Mid-40 LiDAR only, since its price is sufficiently low for sports or surveillance applications.
    We use the following function to simulate its scanning process: 
    \begin{equation}
    \begin{aligned}
       \label{scan}
       r = \alpha \times \cos(3.825\times(\theta_0+0.0017\times n))
    \end{aligned}
    \end{equation}
    where $n \in [0,t\times 1e5]$ and $t$ is in second. $\alpha$ is the scanning radius in pixel, and $\theta_0$ is a random initial angle. This equation is defined in 
    polar coordinate. Final sampling points can be acquired by transforming it from polar coordinates to Cartesian coordinates and translating it to the center of the 
    image space.
 
    As for the avatars, we download various human 3D models from Adobe Mixamo\footnote{https://mixamo.com}. With a view to guarantee the diversity of the generated 
    actions, we can drive these avatars by either ready-made action files or other public datasets' keypoint annotations. Currently, we have developed the driving APIs 
    for COCO17, COCO19 \cite{lin2014microsoft} and SMPL \cite{loper2015smpl} standard keypoint annotated inputs. Regarding the groundtruth matter, we can obtain the 3D human pose from the avatars' humanoid skeleton;
    2D pose label can be acquired via projecting 3D pose into 2D view. Additionally, we can fetch the mesh vertex of each avatar and then compute the semantic segmentation 
    label for each point cloud by considering the pose label meanwhile.

 \noindent\textbf{Unsupervised Domain Adaptation.} \label{self}
    Similar to the 2D projection supervision utilized by other unsupervised or weakly supervised methods \cite{2019self,gholami2022self}, we directly 
    employ off-the-shelf 2D human pose estimation model to get the pseudo 2D pose label $\widetilde{J}_{\text{2D}}$. $\mathcal{L}_{\text{2D}}$ is obtained by calculating the 
    $L_2$ norm between 3D pose projection results and pseudo 2D label from each view.
    \begin{equation}
    \begin{aligned}
       \label{2dloss}
       \mathcal{L}_{\text{2D}} =  \sum_{v=1}^{V}\sum_{k=1}^{K} \left \Vert \mathcal{P}_v(J^{v,k}) - \widetilde{J}_{\text{2D}}^{v,k} \right \Vert_2
    \end{aligned}
    \end{equation}
    where $\mathcal{P}$ is the projection function. $V$ is the number of views. $\mathcal{L}_{\text{2D}}$ is the fundamental loss for unsupervised training.
    In order to attain pseudo 3D label, we choose to use information 
    entropy as one uncertainty index. The entropy of one keypoint prediction heatmap (a spatial probability distribution) $h^k$ is defined as:
    \begin{equation}
    \begin{aligned}
       \label{entropy}
       \mathcal{H}(h^k) = -\sum_{i} h_i^k \times \log h_i^k
    \end{aligned}
    \end{equation}
    where $i$ represents the voxel index. The higher the entropy is, the more uncertain the keypoint location is. 
    To measure a person's uncertainty, we take the maximum entropy value of all keypoints. Different 
    from Kundu et al. \cite{2022uncertainty} who train the uncertainty value, we consider it as an inartificial index. Our experiments demonstrate that 
    the magnitude of entropy values can serve as an indicator of pose estimation quality for a specific network. Therefore, an entropy threshold $\lambda$ can be 
    set to sort out the reasonable predicted 3D human poses. In other words, we select the predicted pose $\widetilde{p}$ whose entropy value is less than $\lambda$ as the pseudo 3D pose 
    label for the next epoch's training, and it can be updated after each training epoch. Therefore, the 3D pseudo pose loss can be obtained by:
    \begin{equation}
    \begin{aligned}
       \label{3dloss}
       \mathcal{L}_{\text{3D}} = \sum_{k=1}^{K} \left \Vert J^k - \widetilde{J}^k \right \Vert_1
    \end{aligned}
    \end{equation}
    where $\widetilde{J}^k$ is the 3D joint of the selected pseudo 3D pose label $\widetilde{p}$.
 
    To further ensure the anatomical plausibility of the pose, we introduce a human prior loss $\mathcal{L}_{\text{prior}}$\footnote{Please refer to Apendix for details.} adapted from Bigalke et al. \cite{bigalke2022domain}. Specifically, we 
    formulate three losses to penalize asymmetric limb lengths $\mathcal{L}_{\text{symm}}$, implausible joint angles $\mathcal{L}_{\text{angle}}$, and implausible bone lengths $\mathcal{L}_{\text{length}}$. 
    To sum up, the final loss function is defined as:
    \begin{equation}
    \begin{aligned}
       \label{loss}
       \mathcal{L}_{\text{unsup}} =\omega_\text{1} \mathcal{L}_{\text{2D}} + \omega_\text{2} \mathbbm{1}(\widetilde{p}<\lambda)\mathcal{L}_{\text{3D}} + \omega_\text{3} \mathcal{L}_{\text{prior}}
    \end{aligned}
    \end{equation}
    where $\omega_\text{1}$, $\omega_\text{2}$ and $\omega_\text{3}$ are the weights of each loss. 

\subsection{Avatar Modeling and Visualization in VR/AR}\label{avatar}
\subsubsection{\textbf{Avatar Modeling}}
In our work, we employ an self-supervised 3D digital avatar modeling method SelfRecon~\cite{jiang2022selfrecon}, and focus on the high-precision reconstruction of players during competitions. The core of our method is the integration of implicit and explicit representations to recover space-time coherent geometries of clothed human bodies from monocular videos. 

Specifically, our modeling process starts with a monocular video that captures a self-rotating player. Through the analysis of video frames, we use explicit representations to capture the player's overall shape, while implicit representations refine the geometric details of clothing and posture. The explicit representation recovers the overall shape through differentiable mask loss, and the implicit details are refined through sampled neural rendering loss and predicted normals. Consistency loss is introduced to maintain coherence between the two geometric representations.

Following SelfRecon, we apply forward deformation fields to generate space-time coherent explicit meshes, which are then updated through non-rigid ray casting and an implicit rendering network to refine the shape of the implicit neural representation. This method allows us to reconstruct high-fidelity, space-time coherent 3D models of players from video captured from a fixed perspective, maintaining high accuracy even when modeling various clothing types.

\subsubsection{\textbf{Visualization in VR/AR}}
In previous sections, we present acquiring player locations and poses, as well as creating digital avatars for each player. In the following, we introduce how these elements are brought to life in virtual environments, allowing audiences to watch the game through VR/AR devices.

Utilizing Unity software, a leading game development platform known for its versatility in creating both 2D and 3D interactive content, we construct a realistic simulation of the sports venue. This digital environment serves as the backdrop for the reconstructed match, offering a virtual stage where the digital avatars can be showcased. The initial step involves importing the digital avatars of the players into the Unity scene. These models are then rigged with skeletal systems, enabling them to mimic the movements and poses of the real players based on the positional and postural data captured during the actual game.

This process of skeletal rigging is crucial, as it translates the athletes' dynamic actions into the digital realm, allowing for a detailed 3D reconstruction of the entire match. By doing so, we create a bridge between the physical performances on the field and their digital counterparts, ensuring that every sprint, tackle, and goal is reflected with precision in the virtual environment.

Furthermore, leveraging Unity's extensive toolset, we implement various interactive features to enhance the spectator experience. These include the ability to rotate the view 360 degrees, enabling viewers to observe the action from any angle. Spectators can also choose their viewing distance and position freely, offering a level of control that goes beyond traditional broadcast views. For a more personalized experience, we introduce a follow-cam feature that allows viewers to adopt the perspective of specific players, providing insight into their movements and strategies throughout the game.

The integration of VR/AR technology with sports visualizations offers a more engaging and immersive experience. Through the use of Unity software and the detailed recreation of player avatars, we are able to bring fans closer to the action than ever before, allowing them to explore the game from within a virtual space that replicates the intensity and excitement of the live event.

\begin{table*}[]\scriptsize
\centering
\caption{Tracking results are evaluated
on BasketBall test set using 3D MOT evaluation tool. Regain module and geometry constraints are denoted as “R” and “G” respectively.}\label{tab:tracking}
\vspace{-0.2cm}
\begin{tabular}{c|cccc|c|c|c|c}
\toprule
& Point cloud & Image & Regain & Geometry & HOTA (\%) & AssA (\%)& MOTA (\%)& IDS \\ \hline
3D &  \checkmark &    &              &                      &  32.86  &  14.69  &  95.24 & 169 \\
2D     &    &  \checkmark &              &                      & 29.42 & 13.08   &  86.71  & 56  \\\hline
2D+3D     &  \checkmark & \checkmark &           &                      &  50.56 &  34.90 & 95.28 &  64 \\
2D+3D R        &  \checkmark & \checkmark  &     \checkmark      &        & 52.91 &  38.03  & 96.14  &  26 \\
2D+3D R+G (\textbf{Ours})     & \checkmark  & \checkmark &  \checkmark &   \checkmark   &  \textbf{55.48}  &  \textbf{41.82}  &  \textbf{96.16}  & \textbf{24}    \\ \bottomrule
\end{tabular}
\end{table*}

\begin{figure}[]
  \centering
  \begin{subfigure}[b]{0.97\columnwidth}
  	\centering
  	\includegraphics[width=\textwidth]{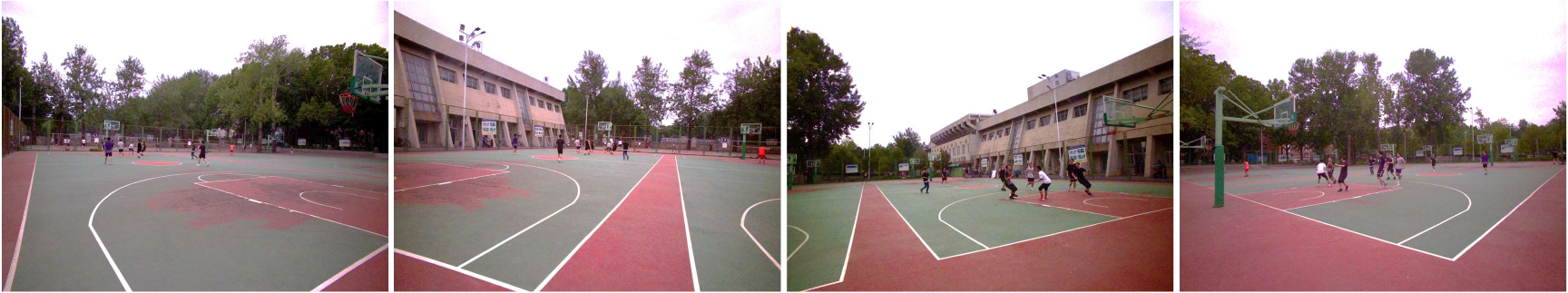}
        \caption{Basketball scenario \uppercase\expandafter{\romannumeral1}.}\label{fig:data_basketball1}
  \end{subfigure}%
  \\
  \begin{subfigure}[b]{0.72\columnwidth}
  	\centering
        \includegraphics[width=\textwidth]{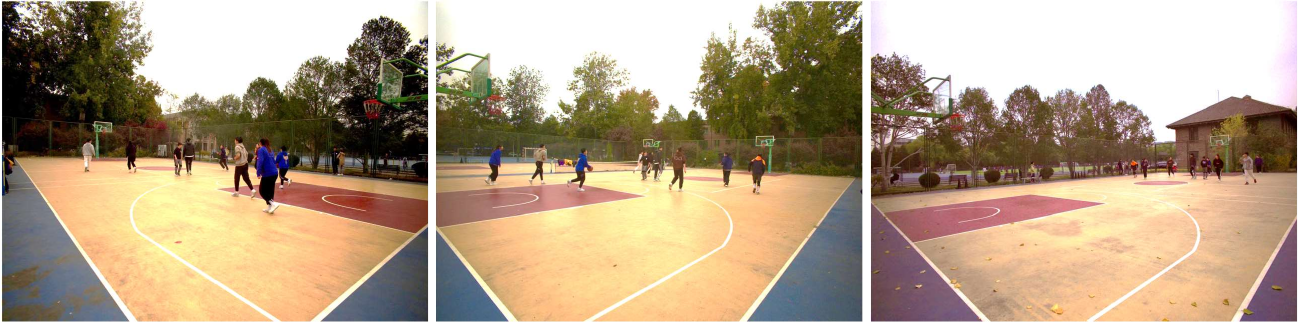}
        \caption{Basketball scenario \uppercase\expandafter{\romannumeral2}.}\label{fig:data_basketball2}
  \end{subfigure}%
  \subfigsCaption{Acquisition scenarios of our dataset}
  \label{fig:scenes}
  \vspace{-.2cm}
\end{figure}

\section{Experiments}
\subsection{BasketBall Dataset}\label{sec:multiplayer}
In this paper, we introduce a \textit{large-scale player tracking dataset}, BasketBall, acquired with a flexible multiview multimodal imaging system, which consists of multiple statically positioned nodes with overlapping fields of view. Each acquisition node is equipped with a Livox Mid-100 LiDAR sensor and a FLIR camera. We synchronize all sensors with $\sim5$ ms accuracy and provide an accurate extrinsic and intrinsic calibration following the method in \cite{3dv}. All the multimodal sequences are recorded with a frame rate of 10 Hz. The images are captured of a resolution of 2048$\times$1536 pixels and the density of LiDAR points is about 30,000 points per frame for each node.

We collect data from two basketball games in different real-world scenarios. A total of nearly 11,000 frames are recorded, which are divided into 3 sequences. The first and second sequences are collected in basketball scenario \uppercase\expandafter{\romannumeral1} by four multimodal nodes, with each part consisting of 2,000 frames. The third sequence comprises 7,000 frames and is collected in basketball scenario \uppercase\expandafter{\romannumeral2} by three nodes. 
As for annotations, we provide 3D bounding boxes and 2D bounding boxes with tracking IDs for 4,000 frames of the first two sequences, consisting of 39,798 3D person detections in total. 



\subsection{Multi-player Tracking}

We evaluate our method on BasketBall dataset, using the first sequence as the training set and the second one as the evaluation set. The 7,000 unlabeled frames are used to verify the reliability of our system. We follow the evaluation setting of ~\cite{DBLP:conf/iros/ab3Dmot} and report the results of MOTA and IDS in CLEAR-MOT metrics~\cite{DBLP:journals/ejivp/clear}. To evaluate association accuracy, we provide the results of HOTA metric~\cite{hota}, which is a more comprehensive metric than CLEAR-MOT to balance the effect of performing accurate detection, association and localization.

\noindent\textbf{Multimodal Fusion Results.}
We use the same 3D detection results provided by our multimodal multiview detector for a fair comparison. Utilizing the same detection results, we compare the tracking results of different modalities. We set $\alpha=0.5$ as a general proportion to fuse two modalities. As shown in Tab.~\ref{tab:tracking}, our multimodal multiview tracking method outperforms the single-modal tracking method by more than $22.62$\% in HOTA. Due to high detection performance, we achieve a bit better performance compared to the single-modal tracking method in CLEAR metrics which overemphasize the detection accuracy. It should be noted that the results in Tab.~\ref{tab:tracking} demonstrate the MOTA metric is not suitable for evaluating long-term object tracking. Our multimodal multiview tracking method achieves a significantly better association accuracy, which is more important for the long-term tracking. This is because our method can better employ the 3D geometry information of point clouds and the appearance information of images to improve the association accuracy. We can observe that tracking only by 3D geometry information performs better than tracking only by appearance information, because the appearance information of images is less robust than the geometry information in 3D space.

\noindent\textbf{Broken Trajectory Regain Module.}
To evaluate the effectiveness of our trajectory regain module, we conduct an ablation study for the data association. We set a plain multimodal tracking method without the regain module and geometry constraints as the baseline. We compare the tracking results of the baseline and our multiview multimodal method with the regain module, evaluated on the BasketBall dataset. As shown in Tab.~\ref{tab:tracking}, the regain module without geometry constraints reduce the ID switches because the regain module can effectively recover the broken trajectories using the appearance feature, which can improve the tracking continuity. Furthermore, our method with geometry constraints outperforms the baseline by $4.92$\% in HOTA. This is because the regain module can effectively recover the broken trajectories using the appearance features, and geometry constraints can ensure reasonable association results. 

\begin{figure}
    \centering
    \includegraphics[width=0.48\textwidth]{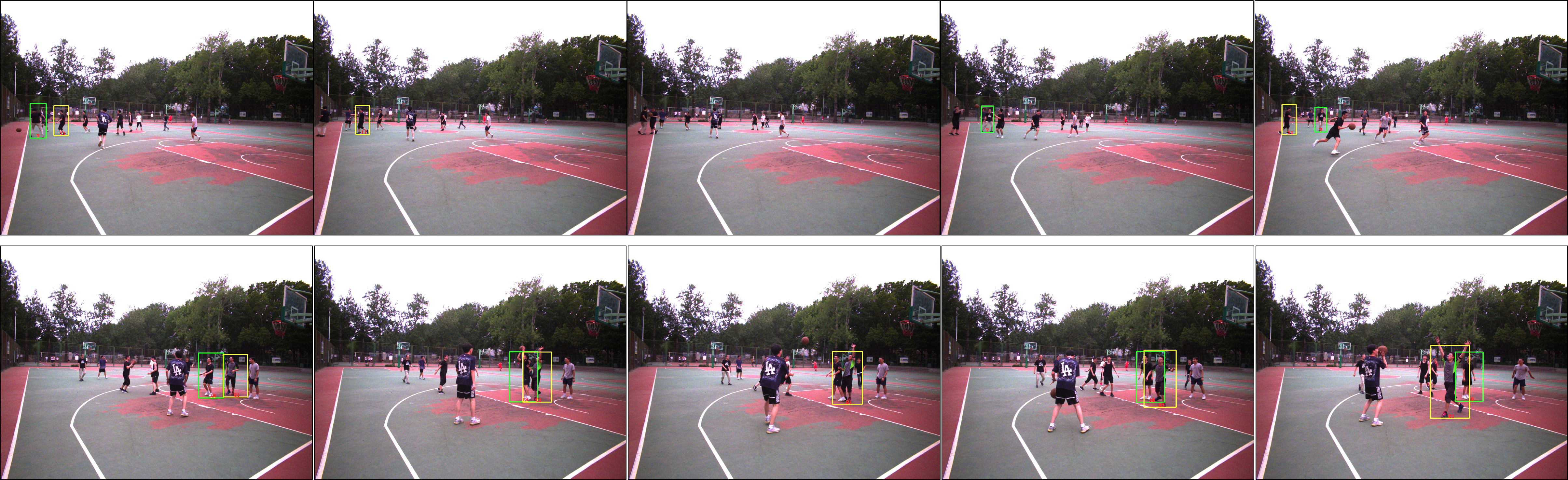}
    \caption{Qualitative results of our tracking method on two special cases: top is the objects leaving the field and bottom is the objects aggregated together. Our regain module can effectively recover the  broken trajectories using the appearance features.}
    \label{fig:cases}
    \vspace{-.2cm}
\end{figure}

We show our qualitative results on two sequences of special cases in Fig.~\ref{fig:cases}. The top sequence is the case where two players leave from the field and reappear, while our broken trajectory regain module is able to retrack the missing objects after they are detected. In the bottom sequence, two players are too close to split them up by detection. The regain module associates the broken trajectory with the detected object again as expected.

\subsection{Pose Estimation}
    In this section, we analyze the performance of our pose estimation pipeline in supervised and 
    unsupervised manner on different datasets. Since the BasketBall dataset does not include annotations for pose estimation, we incorporate CMU Panoptic Studio~\cite{panoptic} to quantitatively evaluate our method. Besides, we also conduct ablation studies to verify the effectiveness of each strategy or loss in unsupervised domain adaption.
 
    \subsubsection{\textbf{Implementation Details}} We use V2V-Net \cite{moon2018v2v} as the point cloud and heatmap voxel's feature extractor. The detailed V2Vnet design is the same 
    with PRN \cite{2020voxelpose}. The resolution of the voxel grid is $64\times64\times64$ defined in a $2 \text{m}\times2 \text{m}\times2 \text{m}$ space. We set $\omega_{\text{1}}=0.02$, $\omega_{\text{2}}=1$, $\omega_{\text{3}}=10$, and $\lambda=6$. We train the voxel-based pose estimation network 
    on an NVIDIA GeForce RTX-3090 (24G) with a batch size of eight. The learning rate is set to 0.001 and the optimizer is Adam \cite{kingma2014adam}.
 
    \subsubsection{\textbf{Datasets and Metrics}} We use the following datasets in our experiments. \\
    \noindent \textbf{I. CMU Panoptic Studio \cite{panoptic}.} It is indoors with a valid scene range around $5 \text{m} \times 5 \text{m}$. In order to realize the multimodal inputs, we select several subsets\footnote{``160422\_ultimatum1", ``160224\_haggling1", ``160226\_haggling1", ``161202\_haggling1", 
    ``160906\_ian1", ``160906\_ian2", ``160906\_ian3", and ``160906\_band1" for training; ``160906\_pizza1", ``160422\_haggling1", ``160906\_ian5", ``160906\_band2" for testing.} that are evaluated in \cite{2021multi} and have depth information. We unify the depth information 
    from Kinect 1 to 5 and process these depth maps by \cref{scan} to get the sparce point cloud which is similar to Mid-40 Livox LiDAR's scanning. About the 2D pose estimation results, we use the results provided by \cite{2021multi} 
    which are claimed to be predicted from HRNet \cite{hrnet}.\\
    \noindent \textbf{II. BasketBall.}  As described in Sec.~\ref{sec:multiplayer}, it has labels for multi-player detection and tracking, while lacks of the 3D keypoints groundtruth. Therefore, it is mainly used for 
    qualitative analysis.\\ 
    \noindent \textbf{III. Player-Sync.} It is a synthetic dataset generated by SyncHuman with the same sensors' configuration and setting as BasketBall. 
     The point cloud is obtained by simulating the scan pattern of Mid-100 Livox LiDARs. It contains ten avatars acting randomly with the first eight 
     for training and the rest two for testing among all frames (3336 frames in 10Hz).
     The predicted 2D human poses are extracted by VitPose \cite{vitpose}. \\
    \noindent \textbf{Evaluation metrics.} For evaluating the 3D human pose, we calculate the standard mean per-joint position error before and after Procrustes Alignment \cite{gower1975generalized} as MPJPE and PA-MPJPE in millimeter respectively.

    \renewcommand\arraystretch{1}
    \begin{table}[t]\scriptsize
       \caption{Comparison of different 3D pose estimation methods on Panoptic and Player-Sync. As to unsupervised manner, we pretrain PointVoxel on Player-Sync in Panoptic testing and on Panoptic in 
       Player-Sync testing.}
       \smallskip
       \label{tab:pose1}
       \centering
       \resizebox{ \linewidth}{!}{
       \begin{tabular}{l c cc cc}
       \toprule
       \multicolumn{1}{c}{\multirow{2}{*}{Methods}} & \multirow{2}{*}{Supervision} & \multicolumn{2}{c}{Panoptic} & \multicolumn{2}{c}{Player-Sync} \\ \cmidrule(lr){3-4} \cmidrule(lr){5-6} 
       \multicolumn{1}{c}{}                                  &                                      & MPJPE            & PA-MPJPE            & MPJPE               & PA-MPJPE               \\ \midrule
       MvP \cite{2021direct}                                                   & sup.                                 & 25.78                 & 25.02                    &  291.11                   &  240.32                      \\
       PlanePose \cite{2021multi}                                             & sup.                                  & 18.54                 & 11.92                    &  119.55                   &  65.80                      \\
       VoxelPose(PRN) \cite{2020voxelpose}                              & sup.                                 & 14.91                        &11.88                    &  40.80                   &  34.34                     \\
       \textit{Ours}                                                   & sup.                                  &  \textbf{14.44}                & \textbf{11.61}                    &  \textbf{31.84}      &  \textbf{27.36}                      \\ \midrule
       DLT \cite{remelli2020lightweight}                               & no sup.                                  & 38.26                 & 44.77                    & 101.49                 & 62.77                        \\
       \textit{Ours}                                                   & no sup.                                  & \textbf{22.00}                 & \textbf{15.96}                    &  \textbf{72.92}                   &  \textbf{62.72}                      \\ \bottomrule
       \end{tabular}}  
    \end{table}

    \renewcommand\arraystretch{1}
    \begin{table}[t]\scriptsize
       \caption{Comparison of different input modalities for PointVoxel on Panoptc and Player-Sync. Model tested on Player-Sync is pretrained 
       on Panoptic, and is pretrained on Player-Sync for Panoptic testing.}
       \smallskip
       \label{tab:pose2}
       \resizebox{ \linewidth}{!}{
       \begin{tabular}{l c cc cc}
       \toprule
       \multicolumn{1}{c}{\multirow{2}{*}{\begin{tabular}[c]{@{}c@{}}Input\\ modality\end{tabular}}} & \multirow{2}{*}{ supervision} & \multicolumn{2}{c}{Panoptic} & \multicolumn{2}{c}{Player-Sync} \\ \cmidrule(lr){3-4} \cmidrule(lr){5-6} 
       \multicolumn{1}{c}{}                                                                          &                              & MPJPE        & PA-MPJPE       & MPJPE          & PA-MPJPE          \\ \midrule
       p.c.                                                                                             & sup.                         & 164.30             & 148.78           &  136.01              & 123.22                  \\
       h.m.                                                                                             & sup.                         & 14.91             & 11.88               &  40.80         & 34.34                  \\
       h.m.+p.c.                                                                                          & sup.                         & \textbf{14.44}             & \textbf{11.61}               &  \textbf{31.84}              & \textbf{27.36}                  \\ \midrule
       h.m.                                                                                             & no sup.                      & 25.72             & 17.24               &  263.05              & 183.85                 \\
       h.m.+p.c.                                                                                          & no sup.                      & \textbf{22.00}             & \textbf{15.96}               &  \textbf{72.92}              & \textbf{62.72}                  \\ \bottomrule
       \end{tabular}}
    \end{table}
    
 \begin{figure}[]
    \centering
    \includegraphics[width=0.7\linewidth]{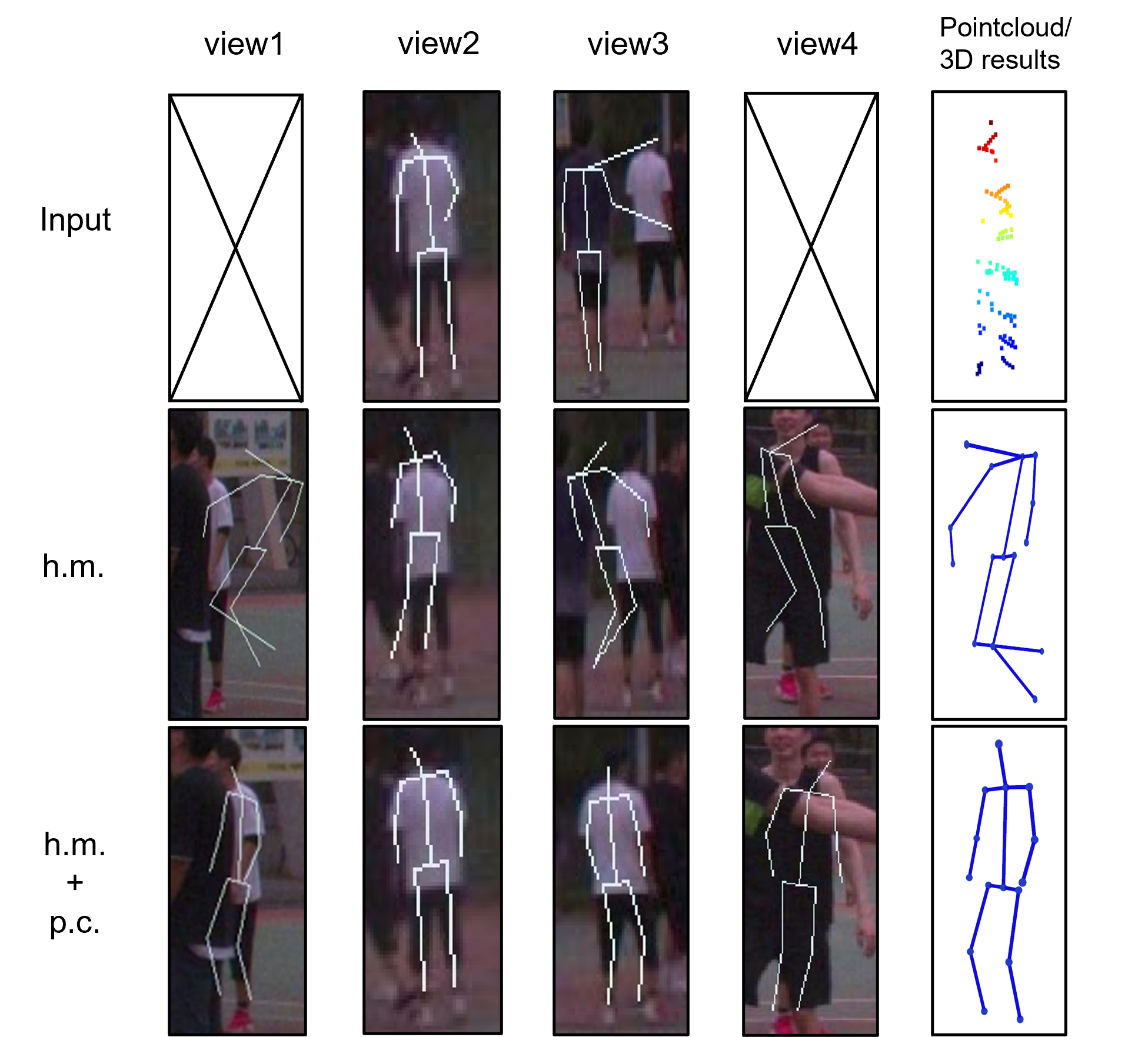}
    \caption{Comparison of different input modalities training on BasketBall. The first row is the 2D pose and point cloud for model inputs, the second row's results are from the model trained by 
    2D pose heatmap (h.m.) inputs, and the third row's results are from the model trained by 2D pose heatmap and point cloud (h.m.+p.c.) inputs. 
    Both models utilize the complete unsupervised training strategy.}
    \label{fig:basketball}
 \end{figure}

 \subsection{3D Pose Estimation Analysis} \label{pose}
    In this section, we evaluate the 3D human pose estimation results of different approaches. For PlaneSweepPose \cite{2021multi}, VoxelPose \cite{2020voxelpose} and PointVoxel in supervised manner, we utilize the groundtruth location of 
    each person into the process of training and testing. As to MvP \cite{2021direct}, it is difficult to add the groundtruth detection information into the network due to its architecture design, which also means that it cannot utilize the depth or point cloud 
    informaion directly. For fairness, we just evaluate the MPJPE@500 metrics (only considering per joint error less than 500 mm) for the matched person as to MvP. 
 
    We evaluate these approaches on Panoptic Studio and Player-Sync as small scene and large scene respectively. Table \ref{tab:pose1} shows that our approach outperforms them in terms of MPJPE and PA-MPJPE especially in large 
    scene's setting. It is worth noting that our approach with unsupervised learning manner can even outperform supervised RGB inputs multiview approaches.
 

    Additionally, we analyze the performance with inputs of different modalities (point cloud, RGB, and both of them) in volumetric architecture. 
    Table \ref{tab:pose2} illustrates that the Livox point cloud information alone
    is unable to accurately extract 3D human skeleton due to its sparsity, and cannot be adopted in unsupervised manner without 2D pseudo labels from heatmaps. 
    However, when combined with RGB information, the performance is greatly enhanced.
    Figure \ref{fig:basketball} demonstrates that multimodal inputs increase the model's tolerance to 2D pose estimation error, including severe situations like miss-prediction or predicting on a wrong person, 
    during unsupervised domain adaption training.


    \begin{table}[]\scriptsize
       \centering
       \caption{Comparison of different unsupervised training strategies. The model is pretrained on Player-Sync for Panoptic testing, and on Panoptic for Player-Sync testing. 
       ``Entropy" means adding entropy selected pseudo 3D pose supervision, and ``Prior" means adding human prior loss.} \label{tab:ablation}
       \smallskip
       \resizebox{\linewidth}{!}{
       \begin{tabular}{c cc cc cc}
       \toprule
        \multicolumn{2}{c}{Unsup. loss} & \multicolumn{2}{c}{Panoptic} & \multicolumn{2}{c}{Player-Sync} \\ \cmidrule(lr){1-2} \cmidrule(lr){3-4} \cmidrule(lr){5-6} 
        Entropy          & Prior         & MPJPE        & PA-MPJPE       & MPJPE          & PA-MPJPE          \\ \midrule
              &          & 29.14             & 16.48               & 84.49               & 75.55                  \\
        \checkmark   &             & 28.26             & 25.85               & 77.38               & 68.84                  \\
              & \checkmark             & 22.88             & 24.69               & 76.45               & 65.54                  \\
        \checkmark      & \checkmark            & \textbf{22.00}             & \textbf{15.96}               &  \textbf{72.92}              & \textbf{62.72}                  \\ \bottomrule
       \end{tabular}}
    \end{table}

 \subsection{Unsupervised Domain Adaption}
    For unsupervised domain adaption, pseudo 2D pose supervision is necessary, and it is the baseline.
    Besides, we adopt an interpretable human prior loss and a pseudo 3D pose (selected by entropy value) loss to assist 
    the learning. Therefore, we conduct an ablation study to analyze the impact of these two losses. 
    From Tab.~\ref{tab:ablation}, entropy selected pseudo 3D pose loss improves the performance, 
    but it does not limit the rationality which means it may generate low-entropy invalid 3D poses. 
    Prior loss controls the results in reasonable action range. 
    Therefore, we choose both losses as our efficient training strategy.  
 
\begin{table}[h]\scriptsize
\centering
\caption{The data statistics of one player in the basketball game.}\label{tab:statistics}
\begin{tabular}{c|c|c|c}
\hline
\hline
\multirow{2}{*}{Player ID} & playing time & running distance/m & sprint distance/m\\
& 3min 18sec & 329.3 & 3.5 \\ \cline{2-4}
\multirow{2}{*}{10} & sprint time/s & jogging time/s & top speed/m/s \\
& 0.5 & 53.5 & 9.6 \\ \hline
\end{tabular}
\end{table}

\begin{figure}[h]
    \centering
    \includegraphics[width=0.32\textwidth]{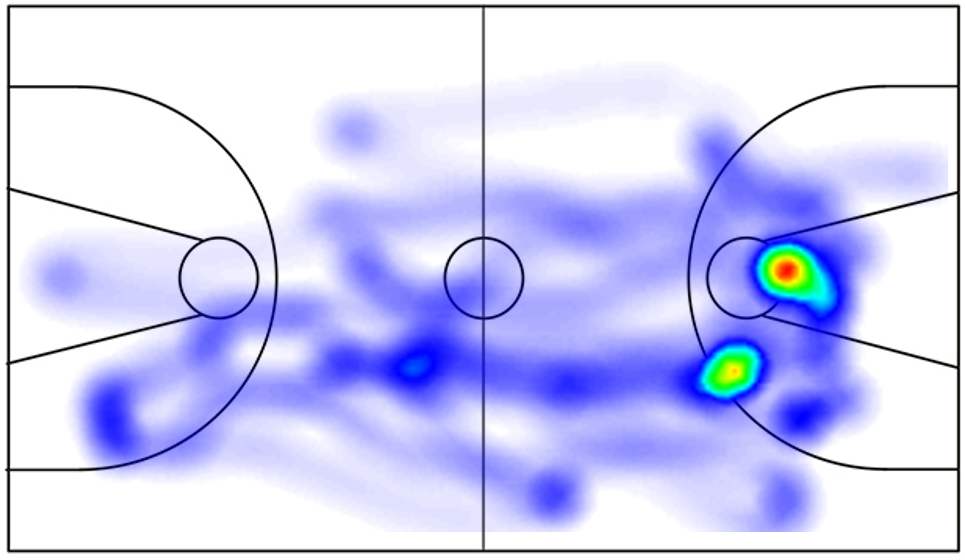}
    \caption{The position heatmap of one player in the game.}
    \label{fig:heatmap}
\end{figure}
\begin{figure}[]
    \centering
    \includegraphics[width=0.48\textwidth]{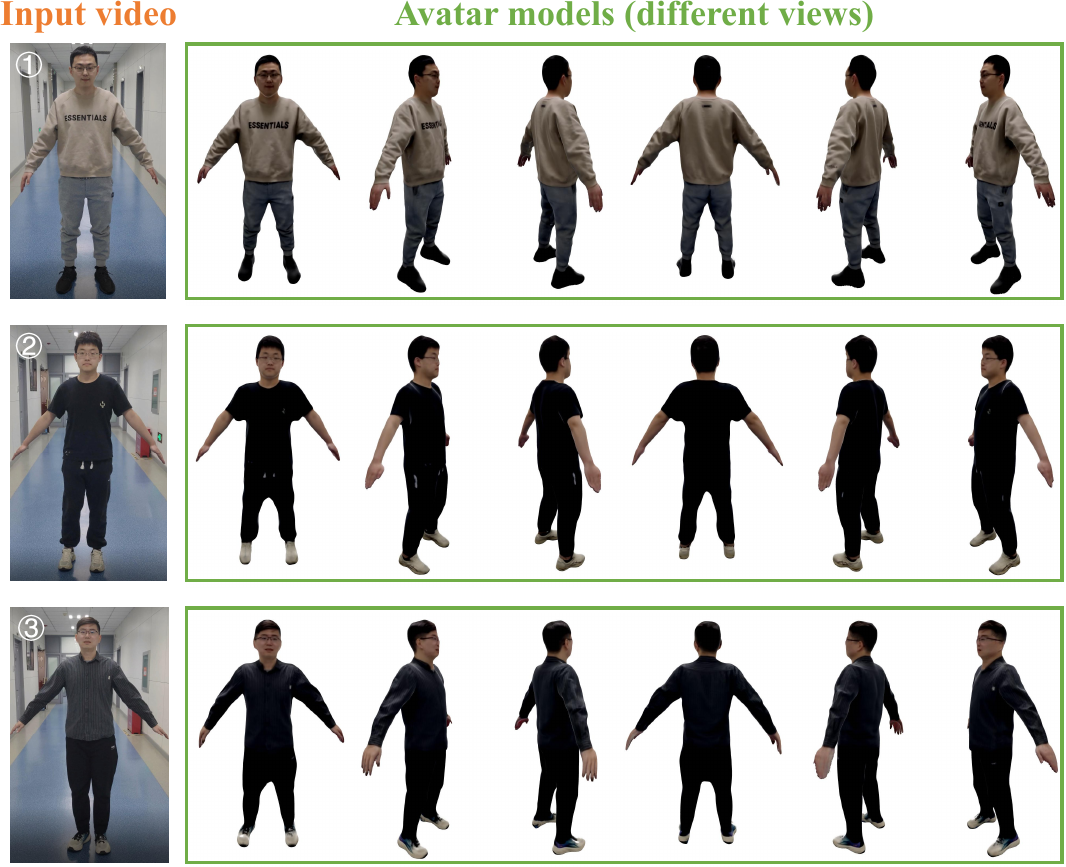}
    \caption{Several examples of our avatar modeling. The left side shows the first frame of the input monocular video with the player rotating in place. The right side displays the results of 3D modeling observed from different viewpoints.}
    \label{fig:avatar}
\end{figure}

\begin{figure}[]
    \centering
    \includegraphics[width=0.48\textwidth]{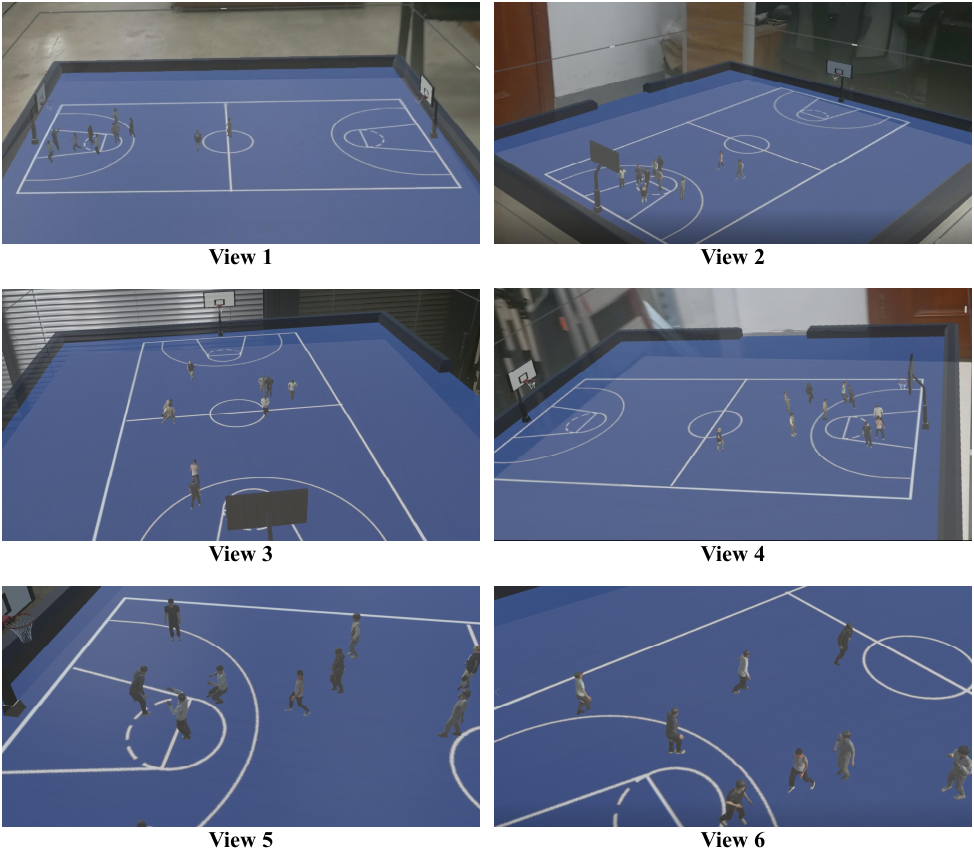}
    \caption{Various perspectives and distances when watching the game with the HoloLens. The six viewpoints are selected from actual game-viewing videos, with more details available in the supplementary material. When watching the game, users can choose any angle and distance they prefer, even opting to follow a specific player.}
    \label{fig:vr}
\end{figure}
\subsection{Sports Analysis}
The player tracking approach described in this paper can be used to build an entire data analysis pipeline. Each player's statistics and analysis can be obtained according to the tracking results, like playing time, position distance, top speed, sprint distance, jogging time, etc. The results of one player are shown in Tab.~\ref{tab:statistics}. In this context, a sprint is defined as a speed of more than 6 m/s and a jog as a speed of less than 1 m/s. Additionally, as shown in Fig.~\ref{fig:heatmap}, we plot a player's position heatmap. The position heatmap helps to analyze the player's movement as well as propensity to assault and defend in certain regions. Besides these, the tracking results can be adopted to complete a wide range of statistics, which is very useful for applications involving game analysis.

\subsection{Avatar Modeling}
With monocular videos, we perform 3D avatar modeling for each player in the match to facilitate visualization on VR/AR platforms. The avatar modeling process results in 3D meshes of the players, with colored textures. These models are drivable, meaning they can be animated or manipulated to replicate the real movements of players during a game. The avatar models of several participants are shown in Fig.~\ref{fig:avatar}. We can observe that these meshes and textures capture the physical appearance, clothing details, and even subtle expressions, contributing to a lifelike representation. This realism enhances the immersive experience for audiences, making them feel like they're witnessing the game firsthand or interacting with real-life athletes in the virtual space.

\subsection{Watching Games in VR/AR}
We drive the avatar models of players based on the results of tracking and pose estimation, visualizing them in virtual space. In this way, audiences can watch the game on VR/AR devices. In Fig.~\ref{fig:vr}, we showcase the effects of watching the game in AR device HoloLens~\cite{HoloLens2024} from different positions and perspectives. The corresponding demonstration video is included in the supplementary materials, presenting an audience using the HoloLens to watch a game. This visualization illustrates the versatility and immersive potential of AR technology in sports viewing, allowing users to experience the game as if they were present in various locations within the stadium or even on the field itself. By adjusting their viewpoint, users can choose their preferred perspective, whether it's a bird's eye view, a sideline perspective, or a position right behind the goalposts. This flexibility enriches the viewing experience, offering a personalized approach to engaging with live sports in a way that traditional broadcasting cannot match. The integration of accurately driven avatar models into these immersive environments further enhances the realism, making the virtual experience more compelling and engaging for fans worldwide.

\section{Conclusion and Future Work}
In conclusion, this work presents a new approach in sports competition analysis and VR/AR visualization. By leveraging mutimodal and multiview LiDARs and cameras for data collection, coupled with a novel framework for multi-player tracking and pose estimation, we enhance the accuracy of sports visualization and analysis. Furthermore, the integration of avatar modeling and actuation of players provides audiences with a more immersive and engaging experience, while also enhancing usability and interactivity. Our system stands out for its real-time performance and cost-efficiency. The extensive experiments underscore the potential of our system to transform the sports domain by providing a more detailed and immersive viewing experience.

While we efficiently complete the 3D modeling and actuation of players in the VR/AR visualization process, we have not accounted for certain props specific to games, such as basketballs and soccer balls in basketball and soccer matches, or racquets in tennis and badminton matches. These props are also important for the audiences' viewing experience. In our future work, we will focus on the modeling and tracking of props used in games. This inclusion aims to enhance the realism and completeness of the virtual viewing experience by accurately representing not only the players but also the essential elements they interact with during a game.

\bibliographystyle{abbrv-doi-hyperref}

\bibliography{template}

\appendix 
\clearpage
\section{Synthetic Dataset Player-Sync}
   We can use our synthetic data generator SyncHuman to simulate any arrangement of sensors to observe a scene. 
   Figure \ref{fig:dataset_sync} shows the datasets we generated, Player-Sync, compared with BasketBall. Using the same scene setting for both training and test yields better transferring performance.

\section{Human Prior Loss in Pose Estimation}
   We use the human prior loss similar to \cite{bigalke2022domain} to encourage the network to generate human-like 3D keypoints. 
   The human prior loss is defined as three parts: 1) the predicting bone length should be in a reasonable range; 2) the predicting length of 
   symmetric bones should be similar; 3) the predicting bone angles should be reasonable according to the kinematic of human.

\begin{figure}
      \centering
      \includegraphics[width=0.35\textwidth]{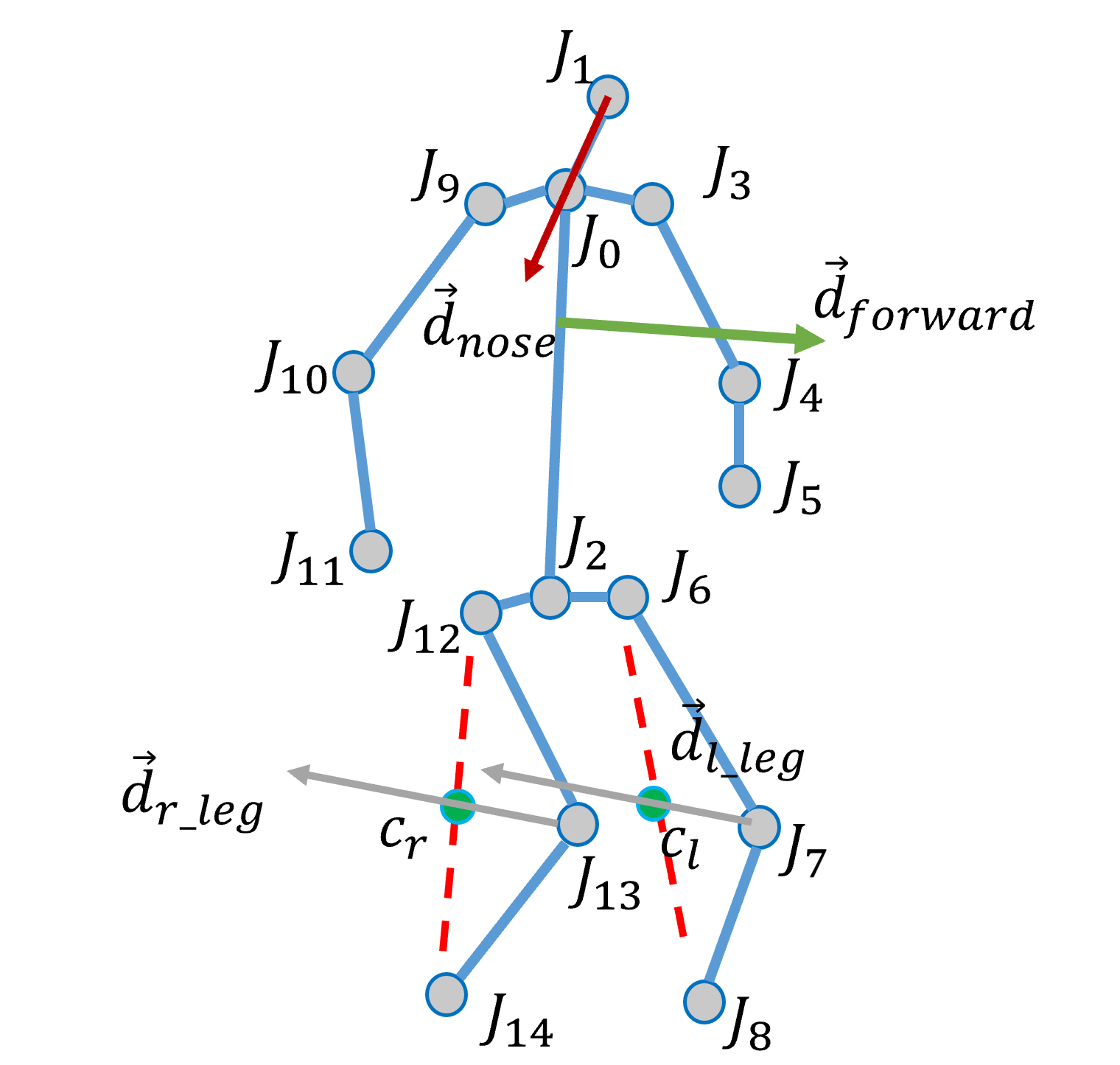}
      \caption{Definition of $\mathcal{L}_{\text{angle}}$.}
      \label{fig:pose angle}   
   \end{figure}

   Different from Bigalke et al. \cite{bigalke2022domain} who distribute each joint with a specific length range, we set one range for all bones. In our case, 
   we set $l_{\text{min}}=0.05 \text{m}$ and $l_{\text{max}}=0.7 \text{m}$. So the $\mathcal{L}_{\text{length}}$ can be defined as:
   \begin{equation}
   \begin{aligned}
      \label{eq:length}
      \mathcal{L}_{\text{length}} = \sum_{b=1}^{N} \mathcal{C}(B_i-l_{\text{max}}, 0) + \mathcal{C}(l_{\text{min}}-B_i, 0) 
   \end{aligned}
   \end{equation}
   where $\mathcal{C}()$ is the clipping function that clip the value greater than $0$, $N$ is the number of bones. 
   As to the symmetric bones, we set the symmetric bones as a pair, and set $L_{\text{2}}$ 
   loss among them. So the $\mathcal{L}_{\text{symm}}$ can be defined as:
   \begin{equation}
   \begin{aligned}
      \label{eq:symm}
      \mathcal{L}_{\text{symm}} = \sum_{b=1}^{N} \Vert B_i - B_{\text{symm}(i)} \Vert_2
   \end{aligned}
   \end{equation}
      
   \noindent where $B_{\text{symm}}$ is the symmetric bone of $B_i$.
   As to angle loss, we limit the nose-neck-midhip angle and hip-knee-ankle angle specifically to let nose be in front of the 
   body and legs be bent forward. Figure \ref{fig:pose angle} shows the definition of each joint and vectors.
   Specially, we do not directly calculate the angle of the bones, but calculate the dot product of corresponding vectors.  
   First, we calculate the forward direction vector $\vec{d}_{\text{forward}}$ of the body, which is the cross product of the unit vector from neck to midhip $\overrightarrow{J_{\text{0}}J_{\text{2}}}$ and unit vector from 
   neck to left shoulder $\overrightarrow{J_{\text{0}}J_{\text{3}}}$:
   \begin{equation}
   \centering
   \label{eq:cross}
   \vec{d}_{\text{forward}} = \overrightarrow{J_{\text{0}}J_{\text{2}}} \times \overrightarrow{J_{\text{0}}J_{\text{3}}}
   \end{equation}
   Then, as to the nose-neck-midhip angle, 
   we calculate the unit vector from neck to nose $\overrightarrow{J_{\text{1}}J_{\text{0}}}$ denoted by $\vec{d}_{\text{nose}}$, 
   and we calculate the dot product of  $\vec{d}_{\text{nose}}$ and $\vec{d}_{\text{forward}}$ and get the head angle loss:
   \begin{equation}
   \begin{aligned}
      \label{eq:head}
      \mathcal{L}_{\text{head\_ang}} = \mathcal{C}(\vec{d}_{\text{forward}} \cdot \vec{d}_{\text{nose}}, 0, 1)
   \end{aligned}
   \end{equation}
   where $\mathcal{C}()$ is the clipping function that clip the value into $0$ to $1$.
   As to the hip-knee-ankle angle, we need to get the midpoint of the hip and ankle denoted by $c_{\text{l}}$ and 
   $c_{\text{r}}$ for left leg and right leg respectively. Then, we calculate the unit vectors from knee point to the leg's midpoint 
   as $\vec{d}_{\text{l\_leg}}$ and $\vec{d}_{\text{r\_leg}}$. Therefore, we get the leg angle loss:
   \begin{equation}
   \begin{aligned}
      \label{eq:leg}
      \mathcal{L}_{\text{leg\_ang}} = \mathcal{C}(\vec{d}_{\text{forward}} \cdot \vec{d}_{\text{l\_leg}}, 0, 1) + 
            \mathcal{C}(\vec{d}_{\text{forward}} \cdot \vec{d}_{\text{r\_leg}}, 0, 1)
   \end{aligned}
   \end{equation}
   where $\mathcal{C}()$ is the clipping function that clip the value into $0$ to $1$.
   Therefore, we can calculate the angle loss:
   \begin{equation}
   \begin{aligned}
      \label{eq:angle}
      \mathcal{L}_{\text{angle}} = \mathcal{L}_{\text{head\_ang}} + \mathcal{L}_{\text{leg\_ang}}
   \end{aligned}
   \end{equation}
   Finally, we combine the three losses together as the human prior loss:
   \begin{equation}
   \begin{aligned}
      \label{eq:bone}
      \mathcal{L}_{\text{prior}} = \gamma_{\text{1}}\mathcal{L}_{\text{length}} + \gamma_{\text{2}}\mathcal{L}_{\text{symm}} + \gamma_{\text{3}}\mathcal{L}_{\text{angle}}
   \end{aligned}
   \end{equation}
   where $\gamma_{\text{1}}$, $\gamma_{\text{2}}$ and $\gamma_{\text{3}}$ are the weights of each loss. In our case, we set 
   all the weights as $1$.

   \begin{figure}
   \centering  
   \includegraphics[width=0.95\linewidth]{Figures/dataset_sync.pdf}
   \caption{Real dataset BasketBall and the synthetic dataset Player-Sync generated by SyncHuman.}
   \label{fig:dataset_sync}
\end{figure}

   \begin{figure}
      \includegraphics[width=0.4\textwidth]{figs/entropy2.png}
      \caption{The entropy value and the specific poses. \includegraphics[height=0.2cm]{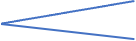} represents an increasing entropy value from left to right among row's samples. 
      The size of joints' ball represents the magnitude of the joint's entropy value.}
      \label{fig:entropy2}
   \end{figure}
      
 \begin{figure}
    \centering
    \includegraphics[width=0.98\linewidth]{figs/ablation_updated.png} 
    \caption{The results show the ablation study about unsupervised training losses on BasketBall. Baseline is only pseudo 2D pose supervision. ``Entropy'' means adding entropy-selected 
    pseudo 3D pose supervision. ``Prior'' means adding human prior loss.}
    \label{fig:basketball2}
 \end{figure}
   \section{Extended Experiments}
   \subsection{Entropy Analysis}
   Figure \ref{fig:entropy2} shows the entropy value and the specific poses, and we can find that the 3D poses become more and more irrational while the entropy goes up.
   
   \subsection{Visualization Analysis on Unsupervised Training Losses}
    For unsupervised domain adaption, pseudo 2D pose supervision is necessary, and it is the baseline.
    Besides, we adopt an interpretable human prior loss and a pseudo 3D pose (selected by entropy value) loss to assist 
    the learning. We conduct a visualization experiment to analyze the impact of these two losses, as shown in Fig.~\ref{fig:basketball2}. We can observe that these losses can enhance the robustness to 2D pose estimation errors. Therefore, we choose both losses as our efficient training strategy. 
\end{document}